\documentclass[conference]{IEEEtran}
\IEEEoverridecommandlockouts
\usepackage{svg}
\usepackage{cite}
\usepackage{amsmath,amssymb,amsfonts}
\usepackage{algorithmic}
\usepackage{graphicx}
\usepackage{textcomp}
\usepackage{url}
\usepackage{booktabs} \usepackage[table]{xcolor}
\def\BibTeX{{\rm B\kern-.05em{\sc i\kern-.025em b}\kern-.08em
    T\kern-.1667em\lower.7ex\hbox{E}\kern-.125emX}}
\begin{document}

\title{Sensitivity Analysis of GRU, LSTM and Transformer Encoder in Classification of Automated Driving  Systems\\
}

\author{\IEEEauthorblockN{Bidhya Shrestha}
\IEEEauthorblockA{
\textit{University of Memphis}\\
bshrstha@memphis.edu}
\and
\IEEEauthorblockN{Christos Papadopoulos}
\IEEEauthorblockA{
\textit{University of Memphis}\\
cppdplos@memphis.edu}
}

\maketitle

\begin{abstract}
Automated driving systems (ADSs) are becoming ubiquitous. Future Software Defined Vehicles (SDVs) may be able to run multiple ADSs, both native and aftermarket such as Comma.ai’s Openpilot. Monitoring systems to independently verify which automated driving system is active are important for safety monitoring, regulatory compliance, insurance assessment, and anomaly detection.

In this paper, we first evaluate the effectiveness of three sequence-based classification models—Gated Recurrent Units (GRU), Long Short-Term Memory (LSTM) networks, and a Transformer encoder model—for identifying Level~2 automated driving systems using vehicle telematics data alone: Comma Openpilot, Tesla Autopilot, and Cadillac Super Cruise, along with manual driving. All three models achieve strong clean-data performance with macro F1-scores of 0.92 (GRU), 0.90 (LSTM), and 0.93 (Transformer encoder model) when trained on clean data; threat-matched training yields 0.904 - 0.916 macro F1 with only a modest clean-data penalty. Second, we introduce a modular robustness evaluation framework that simulates realistic telematics degradation through five corruption families at five severity levels ($L1$ - $L5$). Continuous channels are perturbed using additive white Gaussian noise with cumulative drift, correlated cross-channel noise, and temporal jitter. Binary event signals are subjected to burst loss, delayed transitions, spurious toggles and cross-feature inconsistencies inspired by communication errors. Robustness is measured using macro-F1, which gives equal weight to each class and is suitable for imbalanced multiclass evaluation. Our evaluation reveals a sharp failure-mode split: event-level corruptions reduce macro-F1 only slightly ($\ge$ 0.87 at $L5$), while temporal jitter collapses macro-F1 to ~0.44–0.50 across GRU, LSTM, and Transformer encoder model.
\end{abstract}


\section{Introduction}
\label{newIntro}
The rapid deployment of autonomous and semi-autonomous vehicles has led to a growing diversity of automated driving systems (ADSs), each implementing distinct control algorithms, sensor fusion pipelines, and decision-making strategies. Commercial systems such as Tesla Autopilot~\cite{b1} and Cadillac Super Cruise~\cite{b2} rely on proprietary software stacks, while aftermarket solutions such as Comma.ai’s Openpilot~\cite{b3} further increase software heterogeneity on public roads. Although most widely deployed systems today operate at SAE Level~2 (Partial Driving Automation), higher levels of autonomy are beginning to emerge~\cite{b4}, resulting in vehicles with markedly different control behaviors sharing the same driving environment.


In software-defined vehicles, where multiple ADSs may coexist and be dynamically activated, reliance on self-reported system state is insufficient, since trust in OEM software cannot be assumed. Even under regulatory and cybersecurity compliance regimes, software may malfunction, be compromised, or deliberately misrepresent its operational state. A well-known example is the Volkswagen ``Dieselgate'' scandal~\cite{b5}, in which engine control software was explicitly designed to detect laboratory testing conditions and alter its behavior to pass regulatory checks while violating emissions limits during real-world driving. More recent analyses, including the ``10 Years of Dieselgate'' presentation~\cite{b20}, suggest that similar forms of test detection and behavior manipulation still persist across manufacturers. These findings demonstrate that OEM software can exhibit conditional behavior that diverges from expected operation, underscoring the need for independent, behavior-based monitoring rather than reliance on internal software declarations.

In the context of automated driving, an ADS may misreport whether it is active, operating within its intended design domain, or degraded due to faults or cyber compromise. Such misreporting—whether accidental or adversarial—poses risks to safety, liability attribution, and regulatory enforcement. Independent verification based on observable vehicle behavior therefore becomes essential. Vehicle telematics data, which capture control actions such as steering, braking, and acceleration, provide a practical basis for such verification without requiring access to proprietary software or sensors.

In this work, we build on our prior work~\cite{b9} investigating deep learning–based sequence models for classifying automated driving systems using only vehicle telematics data. We evaluate three representative temporal models: Gated Recurrent Units (GRU)~\cite{b6}, Long Short-Term Memory networks (LSTM)~\cite{b7}, and a Transformer encoder model~\cite{b8}. While GRU is well suited for modeling short-term temporal dependencies and LSTM is well suited for long-term temporal dependencies, Transformer encoder models leverage self-attention to capture both local and long-range patterns in driving behavior. We systematically compare these models in terms of classification performance across multiple ADS platforms.

Beyond classification accuracy, robustness under realistic operational conditions is critical for real-world deployment. Vehicle sensor data are inherently noisy due to sensor noise, calibration drift, timing misalignment, communication latency, packet loss and inconsistencies introduced by distributed vehicle subsystems. To address this limitation, this paper introduces a modular robustness evaluation framework that systematically injects natural telematics corruption into both continuous and event-based vehicle signals that contributed most in the classification of automated driving system.

The proposed framework models continuous sensor degradation using additive White Gaussian Noise (AWGN), stochastic drift processes, correlated cross-channel perturbations, and temporal jitter. In addition, event-level corruption is introduced through communication-inspired perturbation mechanisms including burst loss, delayed state transitions, spurious binary toggles, and cross-feature inconsistencies between logically related signals. Corruption severity is progressively increased to evaluate how sequential models degrade under increasingly adverse operational conditions. Experimental results on $>$ 3.4M labeled samples from~\cite{b9} show that clean data accuracy remains above 90\% for all models, but robustness is corruption-type dependent rather than models dominated. Transformer encoder model retains a small clean-data edge, yet no model is robust to temporal jitter without additional mechanisms.

To our knowledge, this is the first study to introduce a structured robustness benchmark for telematics-based ADS classification that covers both continuous signal corruptions and binary event corruptions across different severity levels. We show that while all models achieve strong clean classification performance, temporal jitter is the dominant failure mode, revealing a key limitation of current sequence models for ADS identification.

The remainder of this paper is organized as follows. Section~II reviews related work; Section~III describes the methodology and datasets; Section~IV presents experimental results; and Section~V concludes the paper.

\section{Related work}
\label{relWork}
Research on ADSs has grown rapidly recent years, driven by advances in vehicle sensors, machine learning and autonomous driving software. Much of the existing work focuses on understanding human driving behavior, driver identification, or system-level performance. However, fewer studies have addressed the classification of different automated driving systems operating under real world conditions. Moreover, the robustness of these models under sensor noise remains largely unexplored. 

Saleh et al.~\cite{b10} proposed a stacked Long Short-Term Memory (LSTM) model to classify human driving behavior into normal, aggressive, and drowsy categories using smartphone sensors which achieved an F1-score of 91\%. However, their approach was limited to smartphone sensor data without incorporating vehicle-based telemetry.

Girma et al.~\cite{b11} developed an LSTM-based driver identification model using vehicle telematics features. They compared LSTM performance against Random Forest, Decision Tree, and fully connected neural networks, showing that LSTM maintained accuracy above 88\% even under noisy conditions, while the other models dropped below 40\%. Nevertheless, their study did not perform feature selection and utilized all 51 available features, indicating that they did not focus on minimizing the number of features. Likewise, they applied noise to random subset of the dataset, making it difficult to interpret the result.

Carvalho et al.~\cite{b12} evaluated the performance of Simple Recurrent Neural Networks (RNNs), LSTMs, and Gated Recurrent Units (GRUs) for driver behavior profiling based on accelerometer data. Their results indicated that GRU achieved the highest accuracy in classifying seven types of driving events, including aggressive braking, turning, and lane changes. However, the dataset was very limited, comprising only 69 samples from four driving trips of roughly 13 minutes each, which restricts the generalizability of their findings.

Guo and Hansen~\cite{b13} demonstrated that a Transformer encoder-based model can outperform LSTM and GRU  for classifying driving behavior into normal, aggressive and drowsy categories, achieving F1-score of 97\%. However, their study was limited by a small dataset of fewer than 10,000 samples.

Shrestha et al.~\cite{b9} showed that GRU models are capable of not only classifying human driving behaviors but also distinguishing between different automated driving systems, including human drivers, achieving an F1-score above 90\%. Nevertheless, their study did not evaluate the robustness of the model under sensor noise, which is critical for real-world deployment where telemetry data is often imperfect. 

These studies highlight the potential of deep learning models for driving system classification but also reveal key gaps: small datasets, limited evaluation across multiple ADS platforms, and a lack of robustness testing. Our work addresses these limitations by applying LSTM, GRU and Transformer encoder model to a large dataset of 3.4M samples across diverse driving scenarios, and evaluates model performance under increasing level of noise and sensor perturbations.
\section{Methodology}
\label{methodology}
\subsection {Problem Formulation}
We formulate ADS identification as a four-class sequential classification problem using only vehicle telematics. Let $x_t \in \mathbb{R}^{11}$ denote the feature vector at time $t$, and let $X_t = [x_{t-T+1}, \ldots, x_t] \in \mathbb{R}^{T \times 11}$ be a fixed-length temporal window with $T = 32$. The model predicts $y_t \in \{0,1,2,3\}$, corresponding to Comma Openpilot, Manual Driving, Tesla Autopilot, and Cadillac Super Cruise, respectively. 

\subsection{Dataset and Features}
We use the telematics dataset from~\cite{b9}, derived from the Comma Dataset~\cite{b14}, Cadillac Dataset~\cite{b15} and Tesla Dataset~\cite{b16}. The final dataset consists of 11 features: six continuous channels \textit{vEgo, aEgo, gas, yawRate, steeringAngleDeg, steeringRateDeg} and five binary event channels \textit{standstill, brakeLightsDEPRECATED, leftBlinker, rightBlinker, brakePressed}. Class labels are encoded as 0 (Comma Openpilot), 1 (Manual Driving), 2 (Tesla Autopilot), and 3 (Cadillac Super Cruise). We use a stratified split of 70\% training, 15\% validation, and 15\% testing with a fixed seed (42).

\subsection{Classification Models}
We evaluate three sequence-learning architectures under a common multiclass classification framework. The GRU model consists of six stacked GRU layers with hidden size 64, followed by a linear classifier applied to the final timestep representation. The LSTM model uses the same overall structure, with six stacked LSTM layers of hidden size 64 and an identical last-timestep classification head. The Transformer encoder model first projects the 11-dimensional input features into a 256-dimensional embedding space, adds a learnable positional embedding over the 32-step input window, and then processes the sequence using six Transformer encoder blocks with eight attention heads. The final timestep representation is passed to a linear classifier for four-class prediction. For clean-data performance analysis, all models are trained using the Adam optimizer with batch size 256, learning rate 0.0001 and a maximum of 100 epochs. Early stopping with patience 7 is applied based on validation loss. To account for class imbalance, class weights are computed from the training labels and incorporated into the cross-entropy loss. Most hyperparameters are selected from commonly used defaults, and no extensive hyperparameter tuning is performed. We evaluate multiple sequence lengths (4, 8, 16, 32 and 64) and learning rates (0.01, 0.001 and 0.0001). Based on validation performance, we select a 32-timestep input window and learning rate 0.0001, although these settings may not be globally optimal.

\subsection{Noise Models}
Table~\ref{table:noise_model} summarizes the telemetry corruption families used in our robustness framework. The corruptions are designed to represent both continuous signal degradation and binary event-level faults observed in realistic vehicle telemetry streams.
\begin{table*}[tbp]
\caption{Noise model used for robustness evaluation}
\begin{center}
\begin{tabular}{|p{3.0cm}|p{4.0cm}|p{4.0cm}|p{5.0cm}|}
\hline
\textbf{Noise Type} & \textbf{Model} & \textbf{Effect} & \textbf{Real-Life Manifestation} \\ 
\hline
Signal stochastic corruption &
Additive Gaussian perturbations at each timestep together with cumulative drift &
Short-term random jitter is superimposed with a slowly growing offset &
Wheel-speed values fluctuate due to vibration and then drift upward as sensors heat or calibration shifts over long drives \\ 
\hline
Signal temporal jitter &
Sequence-wise time misalignment by shifting continuous channels forward/backward &
Breaks cause--effect timing alignment among continuous vehicle dynamics &
Acceleration and steering messages arrive one to two frames late due to ECU clock skew or variable bus scheduling latency \\ 
\hline
Signal correlated corruption &
Joint multivariate perturbation of physically coupled continuous channels &
Related signals are distorted together rather than independently &
CAN gateway fault or shared IMU disturbance simultaneously affects \texttt{yawRate}, \texttt{steeringAngleDeg}, and \texttt{steeringRateDeg}; 

\\ 
\hline
Event-channel corruption &
Burst stickiness, delayed transitions, and spurious toggles in binary signals &
Binary event states may freeze, transition late, or flip unexpectedly &
Turn-indicator state freezes during congestion, brake-light bits flip intermittently from packet errors, or brake transitions appear delayed due to gateway buffering \\ 
\hline
Cross-feature inconsistency &
Logical pair violations between related binary channels &
Creates implausible combinations among dependent event signals &
\texttt{brakePressed=1} with \texttt{brakeLights=0} without hazard context due to message injection, stale cache, or arbitration faults \\ 
\hline
\end{tabular}
\label{table:noise_model}
\end{center}
\vspace{-15pt}
\end{table*}
Each corruption family is evaluated at five severity levels,$L1$ to $L5$, with scale factor $s = L/5$. $L1$ denotes mild perturbation and $L5$ denotes strongest perturbation. Parameter magnitudes increase monotonically with severity within each family. We use family-specific severity mappings rather than a single shared numeric intensity because corruption families operate in different units and mechanisms; this provides comparable semantic difficulty (mild-to-severe) across all families.

 \subsection{Model Training on noise augmented datasets}
 To improve robustness to out-of-distribution telemetry, we use an eval-matched augmentation strategy during training in which the model is intentionally exposed to the same kinds of corruptions during training that it will face at evaluation. For each batch, corruption is applied with probability 0.85; when applied, one corruption family is selected and its severity is sampled uniformly from $L1$ to $L3$.

 \subsection{Benchmark Protocol and Metrics}
 We evaluate each model on: (i) clean test data, and (ii) the full corruption benchmark across all families and severities $(L1–L5)$. For the robustness experiments, we report macro-F1 as the primary metric which gives equal weight to each class and is therefore appropriate for imbalanced multiclass evaluation. This benchmark allows us to measure not only overall robustness but also family-specific degradation behavior and dominant failure modes under progressively harsher telemetry corruption.For clean-dataset performance, we evaluate the models using class-wise precision, recall~\cite{b17}, F1-score and macro-F1~\cite{b18} . We also use confusion matrix to display the distribution of true versus predicted class labels and identify where the model makes errors, highlighting misclassifications between different driving modes.

\section{Results}
\label{results}
We conduct two types of experiments. First, we compare the performance of GRU, LSTM, and Transformer encoder models using clean training data and test data. Second, we evaluate the robustness of these models under mixed telemetry corruptions. Table~\ref{tab:clean_results}, shows that all three deep learning models achieve strong classification performance across the four driving modes: Openpilot, Manual, Autopilot, and Super Cruise. Among the evaluated models, the Transformer encoder model achieves the best overall performance with a macro-average F1-score of 0.93, outperforming both the GRU and LSTM models, which achieve macro-average F1-scores of 0.92 and 0.90, respectively. Across all models, the Super Cruise class consistently achieves the highest precision, recall, and F1-score. In particular, the Transformer encoder model attains a precision of 0.97, recall of 0.99, and F1-score of 0.98 for Super Cruise, indicating that its driving behavior is highly distinguishable from the other classes. Similarly, the GRU and LSTM models also achieve strong performance on this class with F1-scores of 0.97. The Manual driving class is comparatively more difficult to classify for all models. Although the precision values remain relatively high (0.91 to 0.94), the recall values are significantly lower (0.76 to 0.80), indicating that many Manual samples are misclassified as autonomous driving modes. This behavior suggests that certain human driving patterns overlap with autonomous control characteristics, particularly during stable driving conditions.
\begin{table}[tbp]
\caption{Classification Results on Clean Dataset}
\centering
\small
\setlength{\tabcolsep}{3pt}
\resizebox{\columnwidth}{!}{
\begin{tabular}{lccccccccc}
\toprule
& \multicolumn{3}{c}{GRU} 
& \multicolumn{3}{c}{LSTM}
& \multicolumn{3}{c}{Transformer} \\

\cmidrule(lr){2-4}
\cmidrule(lr){5-7}
\cmidrule(lr){8-10}

Class & P & R & F1 & P & R & F1 & P & R & F1 \\
\midrule

Openpilot 
& 0.87 & 0.98 & 0.92
& 0.87 & 0.94 & 0.90
& 0.89 & 0.97 & \textbf{0.93} \\

Manual 
& 0.93 & 0.78 & 0.85
& 0.91 & 0.76 & 0.83
& 0.94 & 0.80 & \textbf{0.86} \\

Autopilot 
& 0.91 & 0.96 & 0.93
& 0.88 & 0.94 & 0.91
& 0.91 & 0.97 & \textbf{0.94} \\

\textbf{Super Cruise}
& \textbf{0.96} & \textbf{0.98} & \textbf{0.97}
& \textbf{0.95} & \textbf{0.98} & \textbf{0.97}
& \textbf{0.97} & \textbf{0.99} & \textbf{0.98} \\

\rowcolor{gray!15}
Macro Avg
& 0.92 & 0.92 & 0.92
& 0.90 & 0.90 & 0.90
& 0.93 & 0.93 & 0.93 \\

\bottomrule
\end{tabular}
}
\label{tab:clean_results}
\vspace{-15pt}
\end{table}
The confusion matrices further support these observations. In the GRU confusion matrix Figure~\ref{fig:cm1}, a substantial number of Manual samples are misclassified as Openpilot (17,465 samples) and Autopilot (12,275 samples). Similar trends are observed in the LSTM confusion matrix Figure~\ref{fig:cm2}, where 17,506 Manual samples are classified as Openpilot and 13,569 as Autopilot. The Transformer encoder model reduces these misclassifications to 14,961 and 11,817, respectively as shown in Figure~\ref{fig:cm3}, demonstrating improved feature representation and temporal modeling capability. For the Openpilot class, the Transformer encoder model again achieves superior performance with an F1-score of 0.93, and correctly classifies 122,968 samples. The GRU and LSTM models correctly classify 123,849 and 119,918 samples, respectively, although LSTM exhibits higher confusion between Openpilot and Autopilot. Similarly, the Transformer encoder model achieves the best performance for the Autopilot class with an F1-score of 0.94 and 135,804 correctly classified samples. The confusion matrices also reveal that misclassifications between autonomous driving modes are relatively limited compared to errors involving the Manual class. For example, the Transformer encoder model misclassifies only 50 Openpilot samples as Autopilot and only one sample as Super Cruise, indicating strong separability between these autonomous systems. Moreover, Super Cruise exhibits the lowest confusion across all models, with the Transformer encoder model correctly classifying 98,679 samples and producing minimal cross-class errors. All three sequence models perform strongly, with the Transformer encoder model achieving the best overall aggregate metrics on clean dataset.

\begin{figure}[htbp]
    \centering
    \includegraphics[width=0.82\columnwidth]{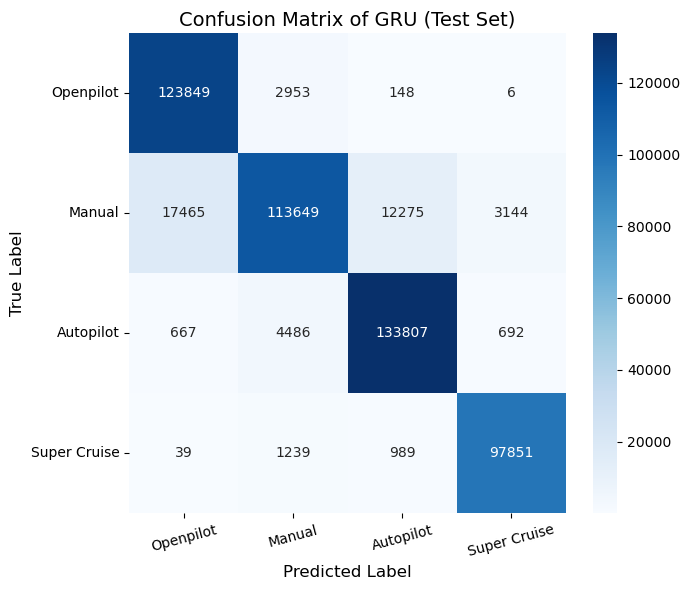}
    \caption{Confusion matrix of GRU}
    \label{fig:cm1}
    \vspace{-15pt}
\end{figure}

\begin{figure}[htbp]
    \centering
    \includegraphics[width=0.82\columnwidth]{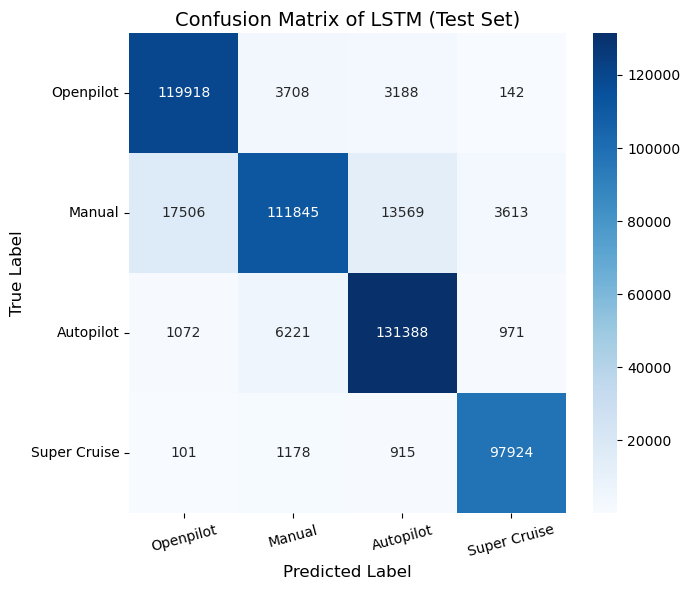}
    \caption{Confusion matrix of LSTM}
    \label{fig:cm2}
    \vspace{-15pt}
\end{figure}

\begin{figure}[htbp]
    \centering
    \includegraphics[width=0.82\columnwidth]{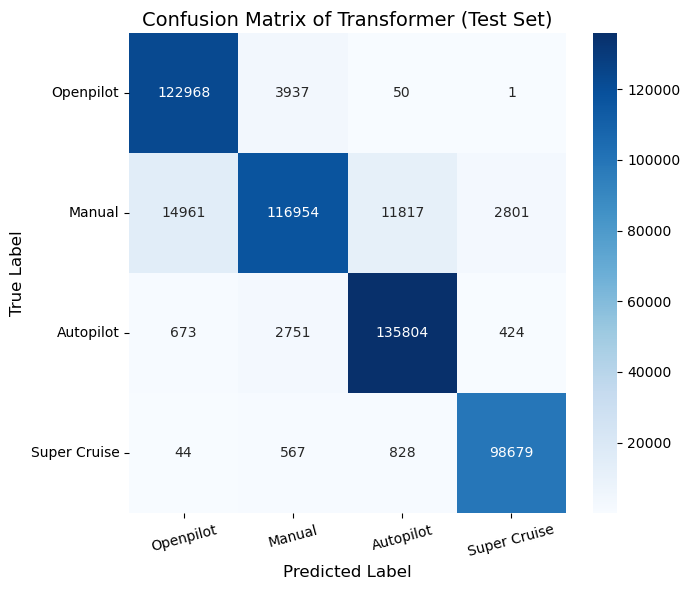}
    \caption{Confusion matrix of Transformer encoder model}
    \label{fig:cm3}
    \vspace{-13pt}
\end{figure}

\subsection{Robustness Testing}
We evaluated the robustness of GRU, LSTM, and Transformer encoder model in presence of noise applied to continuous features (signal-related noise) and noise applied to binary features (event-related noise). 
 Figure~\ref{fig:gru_robustness}, Figure~\ref{fig:lstm_robustness}, Figure~\ref{fig:transformer_robustness} show that the robustness of GRU, LSTM, and Transformer encoder models under (i) signal corruptions: stochastic drift, temporal jitter, and correlated signal perturbation. (ii) event-related corruption: burst/toggle corruption, temporal distortion, and cross-feature inconsistency.
 Under stochastic drift, all three models show gradual degradation as severity increases from $L1$ to $L5$. The GRU macro-F1 decreases from 0.856 at $L1$ to 0.805 at $L5$, while LSTM decreases from 0.855 to 0.810. The Transformer encoder model performs slightly better, decreasing from 0.865 to 0.820. This suggests that all models can tolerate moderate additive noise and drift, but the Transformer encoder model is somewhat more stable under continuous signal perturbations. A likely reason is that self-attention can aggregate information across the full sequence window and reduce dependence on any single noisy timestep. Under correlated signal corruption, the degradation is moderate and relatively stable across severity levels. GRU decreases from 0.797 to 0.780, LSTM from 0.797 to 0.781, and Transformer encoder model from 0.800 to 0.784. Since correlated corruption perturbs physically related channels together, such as speed/acceleration/gas or steering/yaw signals, the models still retain enough structural information to classify the driving system. The relatively small decline indicates that the classifiers learn broader temporal signatures rather than relying on a single continuous feature. The largest weakness appears under temporal jitter. GRU drops from 0.497 at $L1$ to 0.437 at $L5$, LSTM from 0.502 to 0.431, and Transformer encoder model from 0.527 to 0.461. Even though Transformer encoder model is slightly stronger, all models suffer a major collapse. This indicates that ADS classification depends strongly on synchronized timing among continuous channels. 

Under event burst/toggle corruption, performance decreases gradually with severity. GRU macro-F1 falls from 0.901 at $L1$ to 0.869 at $L5$, LSTM from 0.891 to 0.854, and Transformer encoder model from 0.904 to 0.870. This corruption is the most damaging among event-related faults because it can freeze or flip binary states for short intervals. Nevertheless, the decline is much smaller than under signal jitter, suggesting that the classifiers do not depend exclusively on event bits and can compensate using continuous driving dynamics. Under event temporal distortion, macro-F1 remains highly stable across all severity levels. GRU stays around 0.883, LSTM around 0.869–0.871, and Transformer encoder model around 0.878–0.880. This indicates that delayed binary transitions are less harmful than continuous-channel timing shifts. A likely reason is that binary events are sparse and often redundant with continuous dynamics. Under cross-feature inconsistency, all models remain close to clean performance. GRU decreases only from 0.910 at $L1$ to 0.906 at $L5$, LSTM from 0.902 to 0.898, and Transformer encoder model from 0.914 to 0.912. This is the least harmful corruption family.
\begin{figure}[htbp]
    \centering
    \includegraphics[width=1\columnwidth]{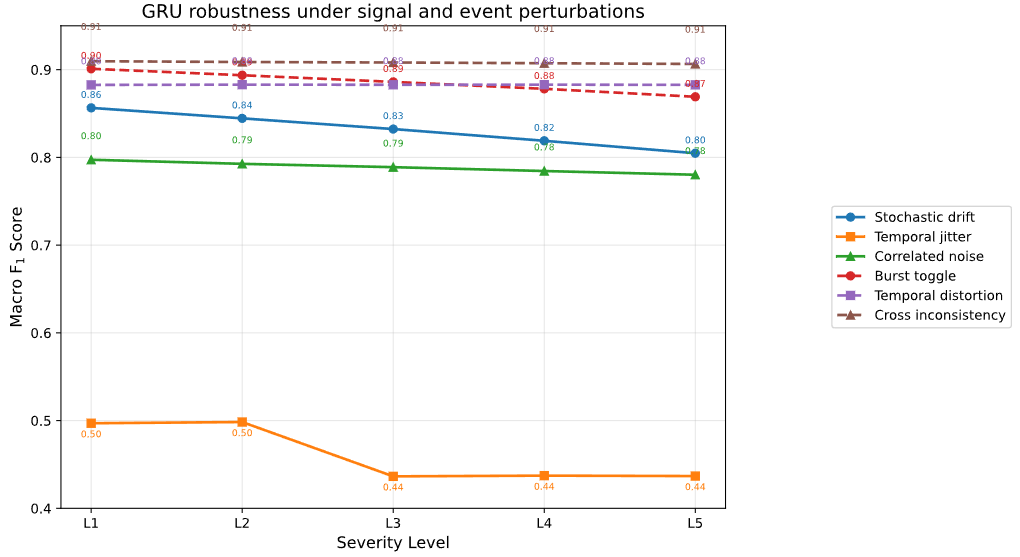}
    \caption{Confusion matrix of GRU}
    \label{fig:cm1}
    \vspace{-15pt}
\end{figure}

\begin{figure}[htbp]
    \centering
    \includegraphics[width=1\columnwidth]{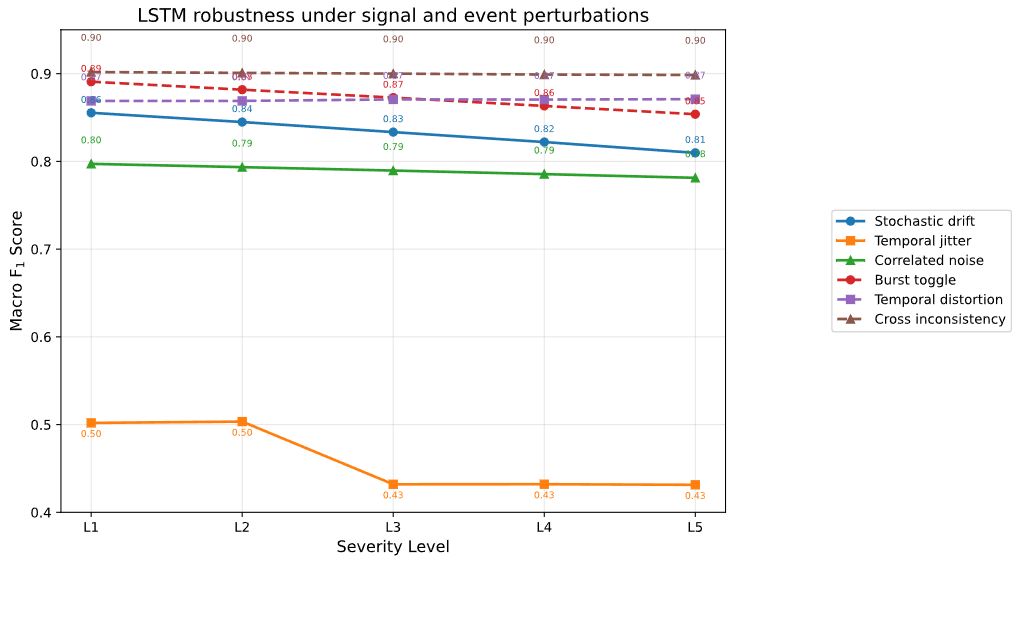}
    \caption{Confusion matrix of LSTM}
    \label{fig:cm2}
    \vspace{-15pt}
\end{figure}

\begin{figure}[htbp]
    \centering
    \includegraphics[width=1\columnwidth]{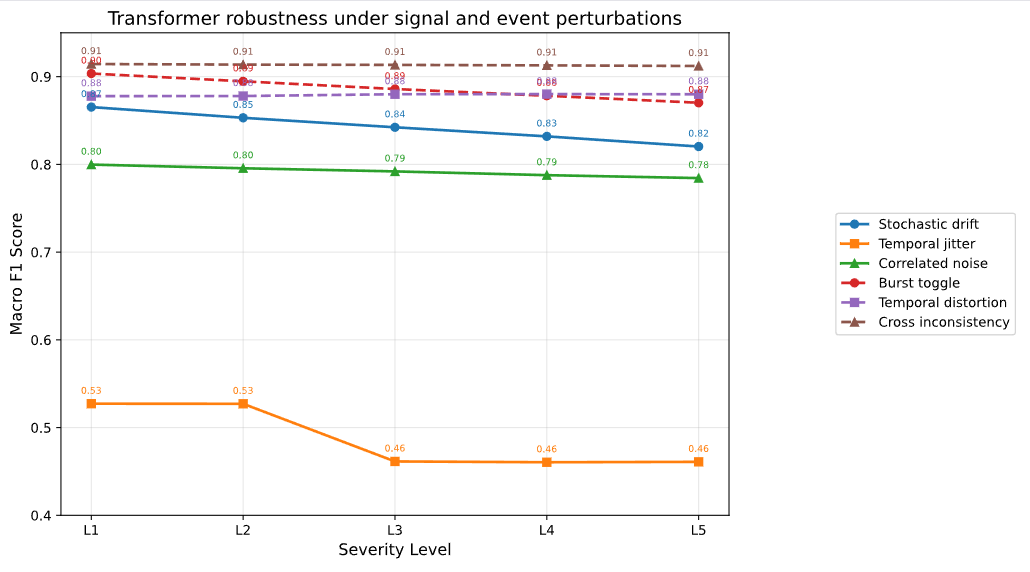}
    \caption{Confusion matrix of Transformer encoder model}
    \label{fig:cm3}
    \vspace{-13pt}
\end{figure}
In summary, event-related corruptions are much less damaging than signal temporal jitter. The Transformer encoder model is again the strongest, especially under cross-feature inconsistency, while GRU is close behind. LSTM is slightly weaker across event corruptions, particularly under burst/toggle faults. These results suggest that the models are generally robust to binary event faults but more vulnerable when the continuous temporal structure of vehicle dynamics is disrupted.

\section{Conclusions}
\label{conclusion}
We evaluate GRU, LSTM, and Transformer encoder models for telematics-only ADS classification across Openpilot, Manual driving, Autopilot, and Super Cruise. All three perform strongly on clean data. It confirms that ADS modes have distinguishable temporal signatures. The Transformer is best overall (macro-F1 0.93), followed by GRU (0.92) and LSTM (0.90), while GRU remains a strong low-complexity option for resource-constrained deployment. Among four classes, Super Cruise is the easiest class, while Manual driving is hardest due to lower recall. Most errors come from Manual samples being predicted as Openpilot or Autopilot, suggesting overlap between stable human driving and automated control in short windows. 

Under mixed corruptions, stochastic drift and correlated perturbations cause gradual degradation from $L1-L5$, with the Transformer generally most stable. Temporal jitter is the dominant failure mode for all models which causes large macro-F1 drops and showing high dependence on cross-channel timing alignment. Event corruptions are much less damaging, with only modest performance loss, indicating that models rely primarily on continuous temporal dynamics and broader context rather than any single binary event channel.

\subsection{Implications and Future Work}
\label{implications}
These findings have direct implications on regulatory compliance and trust in automated driving software. Independent identification of active driving systems based on observable behavior provides a practical mechanism to detect misreporting, unintended mode changes, or deviations from certified operational profiles—whether due to software bugs, misconfiguration, or malicious interference. Such capability is increasingly important as vehicles rely on complex, proprietary software stacks that cannot be directly audited by external stakeholders.

The results show that clean data classification accuracy alone is not sufficient for evaluating automated driving system classifiers. While all three models perform well under clean conditions and remain robust to several event-level faults, timing misalignment in continuous telemetry remains a major vulnerability. For deployment in real-world or security-sensitive environments, robustness evaluation should therefore include structured signal and event corruptions rather than only clean-data metrics. Future work will target temporal-jitter resilience via synchronization-aware preprocessing, jitter-focused augmentation, and time-warp-invariant or hybrid recurrent-attention models~\cite{b19}, and will extend evaluation to additional ADS platforms, longer temporal windows, and broader adversarial threat models.

\end{document}